\pdfoutput=1

\documentclass[11pt]{article}

\usepackage[preprint]{acl}

\usepackage{booktabs}
\usepackage{array}
\usepackage{amsmath} 
\usepackage{multirow} 
\usepackage{times}
\usepackage{latexsym}

\usepackage[T1]{fontenc}

\usepackage[utf8]{inputenc}

\usepackage{microtype}

\usepackage{inconsolata}

\usepackage{graphicx}

\title{Supervised Fine-Tuning Achieve Rapid Task Adaption Via Alternating Attention Head Activation Patterns}

\author{Yang Zhao$^{\spadesuit}\footnotemark[1]$, Li Du $^{\heartsuit}\footnotemark[1]
  $,$\,$ Xiao Ding$^{\spadesuit}\footnotemark[2]$, $\,$ Kai Xiong$^{\spadesuit}$, $\,$  Ting Liu $^{\spadesuit}$and Bing Qin$^{\spadesuit}$ \\
  $^{\spadesuit}$Research Center for Social Computing and Information Retrieval\\ %
  Harbin Institute of Technology, China \\
  $^{\heartsuit}$Beijing Academy of Artificial Intelligence, Beijing, China\\
 \tt{\{yangzhao, xding, kxiong, tliu, qinb\}@ir.hit.edu.cn}\\
 \tt{duli@baai.ac.cn} \,\, \
 \\,\\}

\begin{document}
\maketitle
\renewcommand*{\thefootnote}{\fnsymbol{footnote}}
\footnotetext[1]{These authors contributed equally to this work.}
\renewcommand*{\thefootnote}
{\fnsymbol{footnote}}
\footnotetext[2]{Corresponding Author.}
\renewcommand*{\thefootnote}
{\fnsymbol{footnote}}
\renewcommand*{\thefootnote}
{\arabic{footnote}}
\begin{abstract}
 LLMs' performance on complex tasks is still unsatisfactory. A key issue is that presently LLMs learn in a data-driven schema, while the instructions about these complex tasks are both scarce and hard to collect or construct. On the contrary, a prominent phenomenon is that LLMs can learn rather fast on simpler tasks with adequate prior knowledge captured during pretraining stage. Thus, if the prerequisite and mechanism of such rapid generalization could be elucidated, it could enhance the efficiency and effectiveness of the LLM's ability to learn complex tasks. Thus, in this paper, we employ a gradient-based method, to dissect the process that the SFT process adapts LLMs to downstream tasks via the perspective of attention patterns. We find that: (1) LLMs selectively activate task-specific attention heads during SFT; (2) activation patterns for complex tasks are combinations of basic task patterns; and (3) changes in a few parameters can significantly impact activation patterns after SFT on a small number of samples.
Based on these insights, experiments are conducted to actually enhance the efficiency and effectiveness of SFT.
\end{abstract}

\section{Introduction}

The Supervised Fine-tuning Process (SFT) is essential for optimizing Large Language Models (LLMs) to effectively complete downstream tasks ~\cite{zhang2023instruction}. Leveraging the extensive prior knowledge acquired during pretraining, fine-tuning LLMs on instruction sets enables them to perform well across various tasks. ~\cite{dong2023abilities,xia2024less}. However, in more complex tasks like mathematical reasoning, LLM performance remains unsatisfactory. ~\cite{open-llm-leaderboard-v2,2023opencompass}.
One major reason for such performance limitation is that these complex tasks often require simultaneously using multiple types of knowledge and skills, and thus, more instructions are required to adapt LLMs to these tasks ~\cite{kaplan2020scaling}. However, the more complex the task, the more challenging it is to collect and construct relevant instruction data ~\cite{zhang2023instruction,minaee2024large}. This in turn limits the ability of LLMs to improve performance on these tasks.

\begin{figure} [t]
\small
\includegraphics[width=\linewidth]{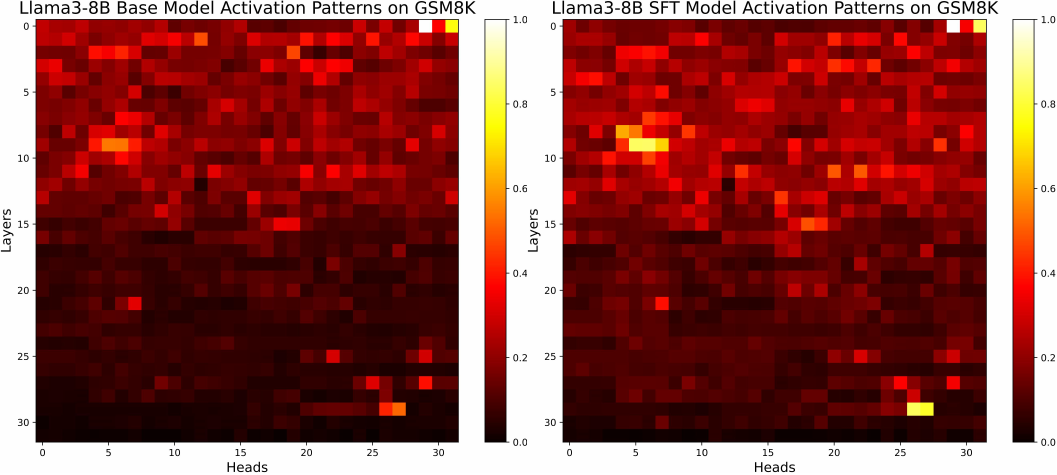}
\caption{Visualization of activation pattern changes in Llama3-8B  on the test set before and after SFT with the GSM8K training set.}
\label{fig:ap}
\end{figure}

In contrast, on most basic tasks, LLMs demonstrate prominent rapid generalization \cite{zhang2023instruction}. By training on just a few thousand instructions, LLMs can learn to complete various tasks\cite{xia2024less}. Therefore, understanding the mechanisms and conditions that enable LLMs' rapid learning and generalization could strongly guide their adaptation to complex tasks.

To address this issue, we propose to analyze the prerequisite and mechanism of such rapid task adaption during SFT from the perspective of \emph{activation patterns of attention heads} using a \emph{gradient-based} method.
Previous studies show that attention heads serve as basic functional units in transformer-based models ~\cite{voita2019analyzing,clark2019does,hao2021self,wang2023label}. These attention heads could capture different types of information and model various relationships for completing different tasks.
Therefore, analyzing how LLMs learn to utilize these basic function units to solve tasks during the SFT process could shed light on the mechanisms behind instruction learning and generalization. If an attention head influences outputs in certain tasks, it is heuristically considered ``\emph{activated}''. The composition of activations across different attention heads forms the \emph{activation pattern} of the LLMs (Figure \ref{fig:ap}), indicating how the model's functional units are integrated to solve a specific task.

To determine if an attention head influences the output, we use a gradient-based method because gradients naturally measure the sensitivity and impact of outputs on input features
 ~\cite{wang2024gradient}. 
By calculating the expected value of the gradient on the attention score matrix, we can quantify the impact of each attention head on model performance before and after SFT
and explains how SFT adapts LLMs to downstream tasks.

From the perspective of attention head activation patterns, we find the critical role of the change of activation pattern in fast learning and generalization during the SFT process: \textbf{(1) LLMs learn to complete tasks by selectively activating attention heads during the SFT process}. After SFT, the LLMs activate certain tasks-specific attention heads, and the differences in activation patterns between tasks become more pronounced; \textbf{(2) LLMs complete tasks by learning to integrate basic skills}.This finding suggests that complex task-solving in LLMs may be decomposed into a series of simpler subtasks, offering a modular perspective on how LLMs can be more effectively fine-tuned for intricate problems;  \textbf{(3) Changes in a few parameters can significantly impact activation patterns after SFT on a small number of samples}. The model's activation patterns exhibit significant changes compared to the base model in the early stage of the SFT process with rather limited optimization steps, showing that minimal parameter adjustments can significantly alter activation patterns.  

We validate our findings by exploring their practical application in enhancing the effectiveness and efficiency of SFT, particularly for complex tasks and in scenarios where high-quality instructions are scarce.
Specifically: (1) When complex task data is limited, by fine-tuning LLMs using instructions on the basic skills required for these tasks, the efficiency of instruction tuning significantly improves. (2) When high-quality domain data is private or unavailable, based on activation patterns, we can select relevant instructions from a large pool of publicly available data to approximate the effects of private data. 
These findings not only validate our understanding of rapid learning and generalization mechanisms but also provide a scalable framework for improving LLM performance in real-world, data-constrained scenarios, such as specialized industries or domains where data availability is limited.

\section{Background and Related Work}

One major factor limiting the learning efficiency of instruction learning of LLMs is the opacity of their internal mechanisms. Though pioneer studies recognized key features or parameters of LLM ~\cite{hao2021self,wang2023label,yang2023idgi}, the mechanism of adapting LLMs towards downstream tasks during SFT is still largely unknown  ~\cite{he2024zero,dong2023abilities}. 
In this paper, we aim to serve as a pioneer to explore these underlying mechanisms.

A critical issue is from which perspective should we start. Instead of focusing on the parameters of LLMs, we choose to investigate the \emph{activation pattern of attention heads}, as: 

(1) Previous studies have identified that an attention head can act as a basic functional unit in transformer-based models ~\cite{voita2019analyzing,clark2019does,wang2023label}, with different attention heads serving distinct functions. For example, ~\citet{voita2019analyzing} and ~\citet{clark2019does} found that attention heads at different layers focus on different parts-of-speeches, while ~\citet{wang2023label} observed that during in-context learning, different attention heads focus on different parts of the prompts.

(2) Additionally, prior research ~\cite{jin2024cutting,lee2024automatic} has suggested the task-specific activation of attention heads, as pruning certain attention heads would lead to task-specific performance degradation. Heuristically, since a task can be viewed as a combination of several basic tasks, completing it requires the coordination of multiple corresponding basic functional units, i.e., attention heads. 

(3) In contrast, Previous studies \cite{yu2024language,fu2023effectiveness} have suggested that parameter changes after SFT are limited. Furthermore, given the vast number of parameters in LLMs, interpreting the significance of these parameters and their changes are highly challenging.

For clarity, we define attention heads that impact downstream tasks as \emph{activated} and those that do not as \emph{inactive}.
The matrix representing the influence of attention heads on downstream tasks is called the \emph{activation pattern}. 


\section{Methodology}

During pretraining, LLMs learn to use different attention heads to capture various types of information  ~\cite{voita2019analyzing, khaki2024need}. Heuristically, a complex task can often be viewed as a composition of several basic tasks. For instance, solving a mathematical problem using code, can be broken down into fundamental tasks like code generation and mathematical reasoning. Since each attention head may correspond to a specific function, an LLM can invoke multiple attention heads and integrate their functions to accomplish a complex task. Combining these observations leads to our core assumption: \textbf{ during the SFT process, LLMs quickly adapt to downstream tasks by invoking different attention heads.}
Thus, in this paper, we investigate the mechanism of fast generalization in the SFT process by analyzing changes in the activation patterns of attention heads.

\subsection{Gradient-Based Analysis of Attention Head Activation Patterns}
A key issue is determining whether an attention head is \emph{activated} during a task. Inspired by \cite{hao2021self} and \cite{wang2023label}, we measure this using model gradients, as they reflect the influence of input features or parameters on the model's output ~\cite{wang2024gradient}. A larger gradient for a particular attention head suggests that the model's output is more sensitive to it, indicating a stronger influence on outputs.

Specifically, given an LLM with~$L$ layers and~$H$ attention heads per layer, and a dataset $\mathcal{T}$, the activation pattern~$AP^{\mathcal{T}}$ can be represented as an~\(L \times H\) matrix, with each element representing the activation level reflecting the contribution of one specific attention head to a given task:
\begin{equation}
\label{AP}
\setlength{\abovedisplayskip}{0pt}
\setlength{\belowdisplayskip}{0pt}
AP^{\mathcal{T}} = \{ AL_{l,h}^{\mathcal{T}} \}_{l\in[1,\dots,L], h\in[1,\dots,H]} ,
\end{equation}
where each element~\(AL_{l,h}\) corresponds to the activation level of the \(h\)-th attention head in the \(l\)-th layer during the given task, which is measured as:
\begin{equation}
\label{AL}
\setlength{\abovedisplayskip}{0pt}
\setlength{\belowdisplayskip}{0pt}
AL_{l,h} = \frac{1}{N}\sum_i {\Gamma_{l,h}^T}\frac{\partial L(x_i)}{\partial \Gamma_{l,h}},
\end{equation}
where \(N\) represents the size of $\mathcal{T}$, \(x_i\) is the \(i\)th instance in $\mathcal{T}$ and \(L(x_i)\) is the loss value of the model for \(x_i\). \(\Gamma_{l,h}\) is the attention matrix of the $h$ th attention head in the $l$ th layer.\(\frac{\partial L(x_i)}{\partial \Gamma_{l,h}}\) is the gradient of the $h$ th attention head in the $l$th layer with respect to \(L(x_i)\). By combining the absolute values of attention scores with the sensitivity of the loss to these scores, the total influence of a specific attention head on outputs and the task can be quantified. Thus, $AP^{task}$, composed of $AL^{task}$, reflects how each attention head contributes to task completion.

\subsection{Analytical Framework}
To validate our core assumption, we focus on three progressive issues: (1) How do activation patterns change during the SFT process for each task, and are these changes task-specific? In other words, can we grasp the characteristics of the task from the perspective of activation patterns?
(2) What is the relationship between activation patterns in complex tasks and basic tasks? This reflects the prerequisites that may be necessary for learning complex tasks; and (3) on basic tasks, how many training samples are required to significantly change the activation patterns? The necessary sample size indicates the model's ability to achieve rapid generalization.

First, we evaluated the changes in activation patterns across tasks before and after SFT. We observed that more attention heads are activated after SFT, and these activations are task-specific, suggesting that attention heads are selectively activated during SFT.
Next, we explored the relationship between activation patterns in complex and basic tasks, finding that the patterns of several basic tasks effectively approximate those of complex tasks. This led us to conclude that activation patterns in complex tasks are combinations of basic task patterns.
Finally, dynamic observations during SFT showed that even small datasets can significantly alter activation patterns.

To further validate our conclusions, we applied these insights to enhance SFT effectiveness on complex tasks. The resulting improvements further confirmed our theory. The following sections provide a detailed analysis and conclusions.

\section{SFT Adapts to Downstream Tasks by Modifying Activation Patterns}

This section uses gradient-based analysis to examine the effect of SFT on transformer attention head activation patterns. We identify three key insights: (1) After SFT, LLMs accomplish tasks by selectively activating specific attention heads. (2) Changes in activation patterns for compound tasks can be interpreted as a combination of those in more basic tasks. (3) During SFT, activation patterns exhibit swift changes, even small datasets can cause significant changes.

\subsection{Attention Heads are Selectively Activated}
\label{4.1}
In this section, we analyze changes in attention head activation patterns before and after SFT across various tasks and explore the relationships between these changes.

\subsubsection{Analytical Method}
To analyze changes in model activation patterns before and after SFT for each task, we used three metrics: Gini Coefficient, Coefficient of Variation (CV), and Kurtosis. The Gini Coefficient measures distribution inequality, with values above 0.4 indicating a concentration of activation in a few attention heads. The Coefficient of Variation reflects relative dispersion, with values over 1 signaling substantial variability across attention heads. Kurtosis describes the distribution's shape, where values above 3 suggest heavy tails and sharp peaks, indicating the presence of extreme values


\subsubsection{Experiments Setup}
We conducted experiments on models with varying performance levels, including Llama3-8B ~\cite{llama3modelcard}, Gemma-7B ~\cite{team2024gemma}, and OPT-6.7B ~\cite{zhang2022opt}, across a range of tasks: mathematical reasoning (MATH ~\cite{hendrycksmath2021}, GSM8K ~\cite{cobbe2021gsm8k}), coding (Code Search Net ~\cite{husain2019codesearchnet}, SGSM ~\cite{christ2024mathwell}), and natural language processing and reasoning (HellaSwag ~\cite{zellers2019hellaswag}, Winogrande ~\cite{sakaguchi2021winogrande}, ARC ~\cite{clark2018think}). 

\subsubsection{Results}
\begin{table}[t]
\centering
\small
\setlength{\belowcaptionskip}{-1pt}
\renewcommand{\arraystretch}{1}
\begin{tabular}{
    >{\centering\arraybackslash}m{1,5cm}
    >{\centering\arraybackslash}m{1.5cm}
    >{\centering\arraybackslash}m{0.9cm}
    >{\centering\arraybackslash}m{0.9cm}
    >{\centering\arraybackslash}m{0.9cm}
}
\toprule
Model & State & Gini  & CV & Kurtosis \\ 
\midrule
\multirow{2}{*}{Llama3-8B} & Before SFT & 0.50 & 1.19 & 95.37 \\ 
 & After SFT & 0.33 & 0.71 & 39.55 \\ 
\multirow{2}{*}{Gemma-7B} & Before SFT & 0.48 & 1.71 & 240.99 \\ 
 & After SFT & 0.38 & 1.23 & 130.82 \\ 
\multirow{2}{*}{OPT-6.7B} & Before SFT & 0.42 & 1.06 & 50.67 \\ 
 & After SFT & 0.38 & 0.819 & 24.23 \\ 
\bottomrule
\end{tabular}
\caption{Statistics on the distribution of activation patterns for different LLMs. Experiments were conducted on tasks such as Code Search Net, GSM8k, MATH, SGSM, ARC, HellaSwag, and Winogrande.}
\label{tab:mul-head}
\end{table}

As shown in Table~\ref{tab:mul-head}, before SFT, base models exhibited significant outliers (Kurtosis > 3), uneven distribution (Gini > 0.4), and high variability (CV > 1) in activation patterns, suggesting that only a few attention heads contributed to the tasks. After SFT, activation patterns became more uniform, with decreases in Gini, CV, and Kurtosis, indicating that SFT adapts LLMs by increasing activation levels across attention heads. However, activation values remain skewed, showing that a few heads still dominate.


\begin{figure} [t]
\small
\includegraphics[width=\linewidth]{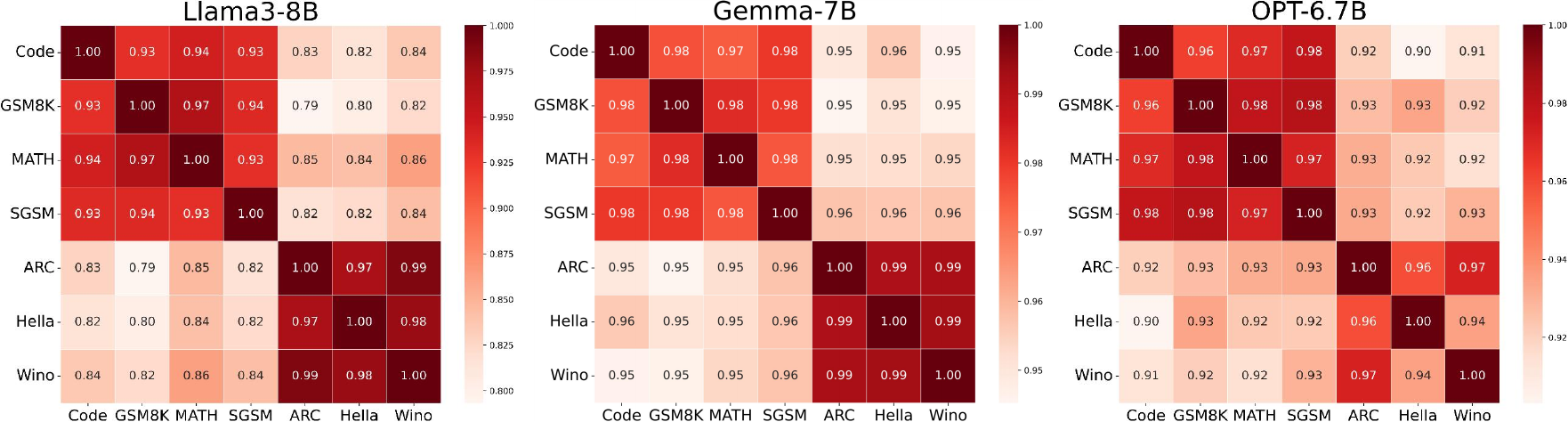}
\small
\includegraphics[width=\linewidth]{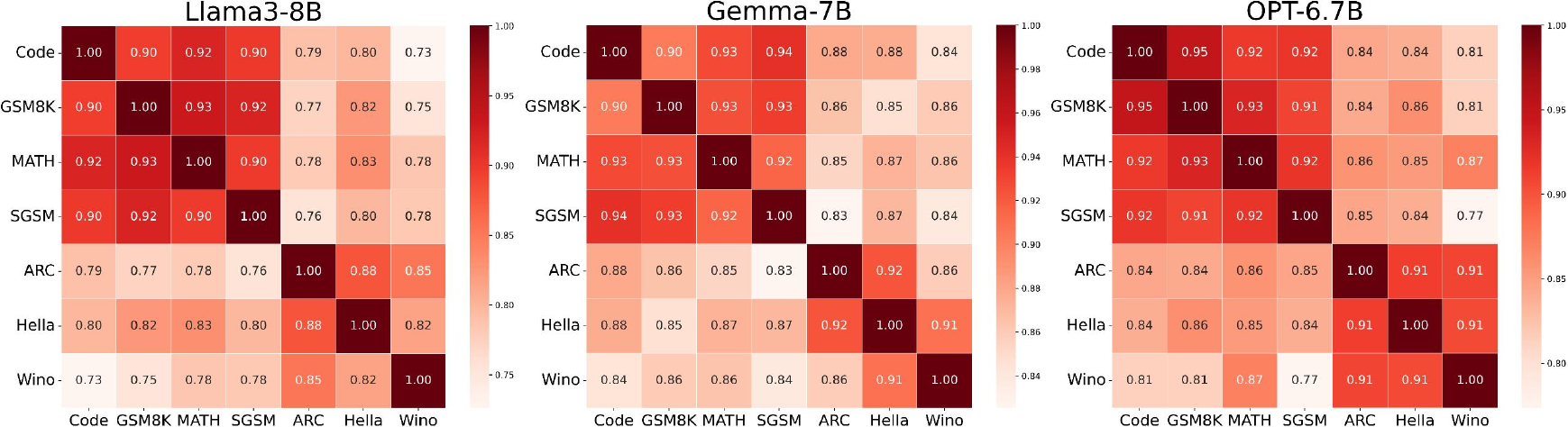}
\small
\includegraphics[width=\linewidth]{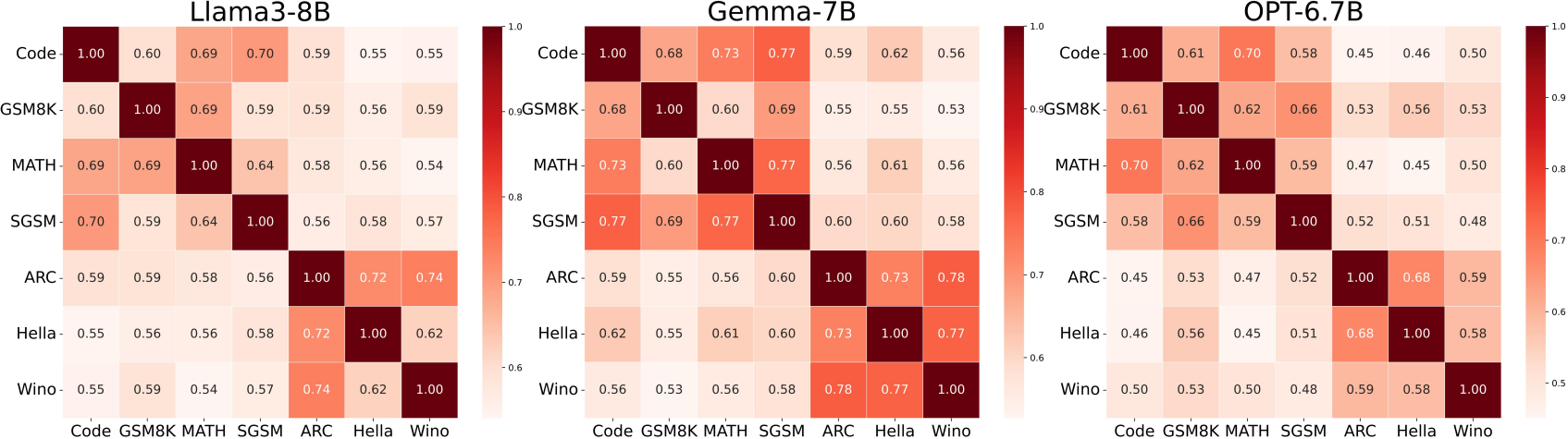}

\caption{The correlation coefficients of the activation pattern change rates for the Llama3-8B, Gemma-7B, and OPT-6.7B models on tasks  before SFT, after SFT, and during the SFT process (corresponding to the top, middle, and bottom sections, respectively).}
\label{fig:multi-task_activation_patterns}
\end{figure}

Figure~\ref{fig:multi-task_activation_patterns} shows the similarity in activation pattern change rates across tasks for three models before and after SFT, measured by correlation coefficients. After SFT, tasks grouped into two categories—math/code and text reasoning—aligning with human understanding of task relationships. This suggests that attention patterns reflect task characteristics and specificity. Post-SFT, stronger task specificity is observed, as indicated by decreased correlation coefficients between tasks, implying reduced similarity.


Considering that the activation levels increase after SFT, we deduce that SFT adapts models to specific tasks by selectively enhancing task-relevant activation patterns.

\subsection{Activation Patterns on Complex Tasks are Combinations of Basic Tasks}
\label{4.2}
This section focuses on the relationship between activation pattern changes in basic and complex tasks. Understanding this relationship is key to equipping models with the necessary prerequisites for rapid learning of complex tasks. 

\subsubsection{Analytical Method}

\begin{figure} [t]
\small
\includegraphics[width=\linewidth]{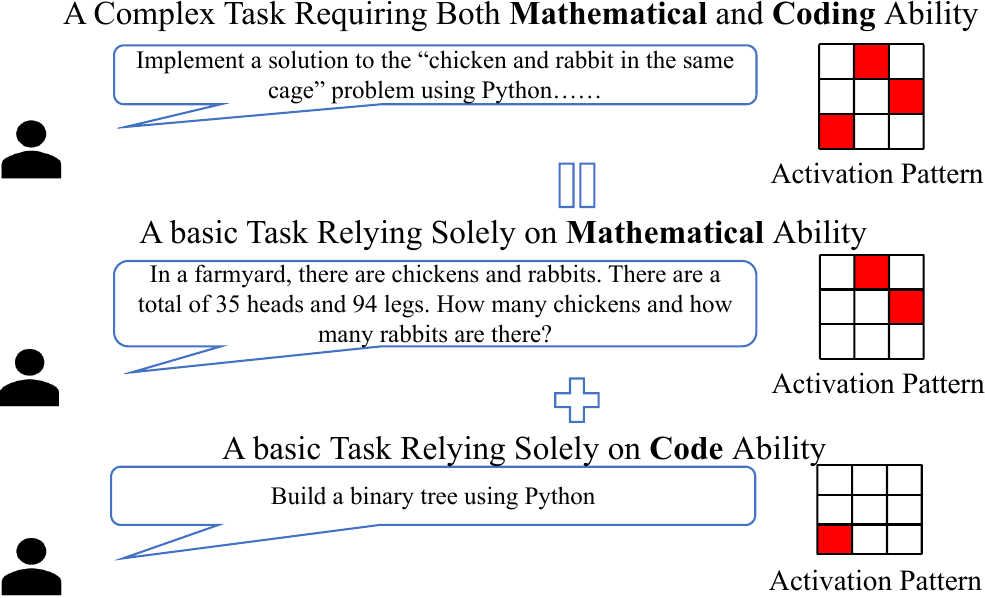}
\caption{The two tasks above are basic tasks that each rely on a single skill, while the one below is a complex task that relies on both coding and mathematical skills.}
\label{pic:task}
\end{figure}

In this paper, we define \emph{complex tasks} as those that can be decomposed into basic skills. For example, as shown in Figure~\ref{pic:task}, solving the  ``chicken and rabbit'' problem requires mathematical reasoning, while constructing a binary tree requires Python coding—both of which are basic tasks. However, solving the ``chicken and rabbit'' problem using Python code is a complex task, as it integrates both mathematical reasoning and coding skills.

The actual relationship between activation patterns in basic tasks and complex tasks can be quite intricate. Without generality, in our study, we employ a linear function to model the relationship between the activation patterns of basic and complex tasks. 
Specifically, we take the changes in activation patterns for complex tasks as the dependent variable, and employ the changes in activation patterns for multiple simple tasks as independent variables, using the least squares method to fit the regression coefficients.
\begin{equation}
\setlength{\abovedisplayskip}{-1pt}
\setlength{\belowdisplayskip}{-1pt}
\label{ercheng}
\Delta AP^{complex} = \sum_{i=1}^{n} \alpha_i \Delta AP^{basic_i} + \epsilon ,
\end{equation}
where \(\alpha_i\) is a regression coefficient representing the contribution of each basic task \(i\) to the change in activation patterns for the complex task, \(n\) is the total number of basic tasks, and \(\epsilon\) is the error term.
Subsequently, we employ the R-squared (\(R^2\)) value to measure the goodness of fit, which represents the proportion of the variance in the dependent variable that can be explained by the independent variables.

\subsubsection{Experiments Setup}

In this study, we conducted two groups of experiments: (1)  We take MATH, GSM8K, Code Search Net, HellaSwag, Winogrande, and ARC as the basic tasks, and select the SGSM task, which requires coding ability and elementary math skills, as the complex task.
(2) We conduct experiments on the Infinity Instruct dataset \cite{zhao2024iidoptimizinginstructionlearning}, which provides each instruction tags describing the skills and knowledge required to complete that instruction, so that we can select both basic and complex task-related instructions. Specifically, we selected instructions requiring only a single skill, including `mathematics', `logical reasoning', `programming and software development', `education and consulting', `financial and business knowledge', and `legal knowledge', to form into the basic task data set. Next, we selected instructions requiring both `logical reasoning' and `programming/software development' skills as the complex tasks for SFT and calculated their activation patterns.

\subsubsection{Result}

\begin{figure} [t]
\small
\includegraphics[width=\linewidth]{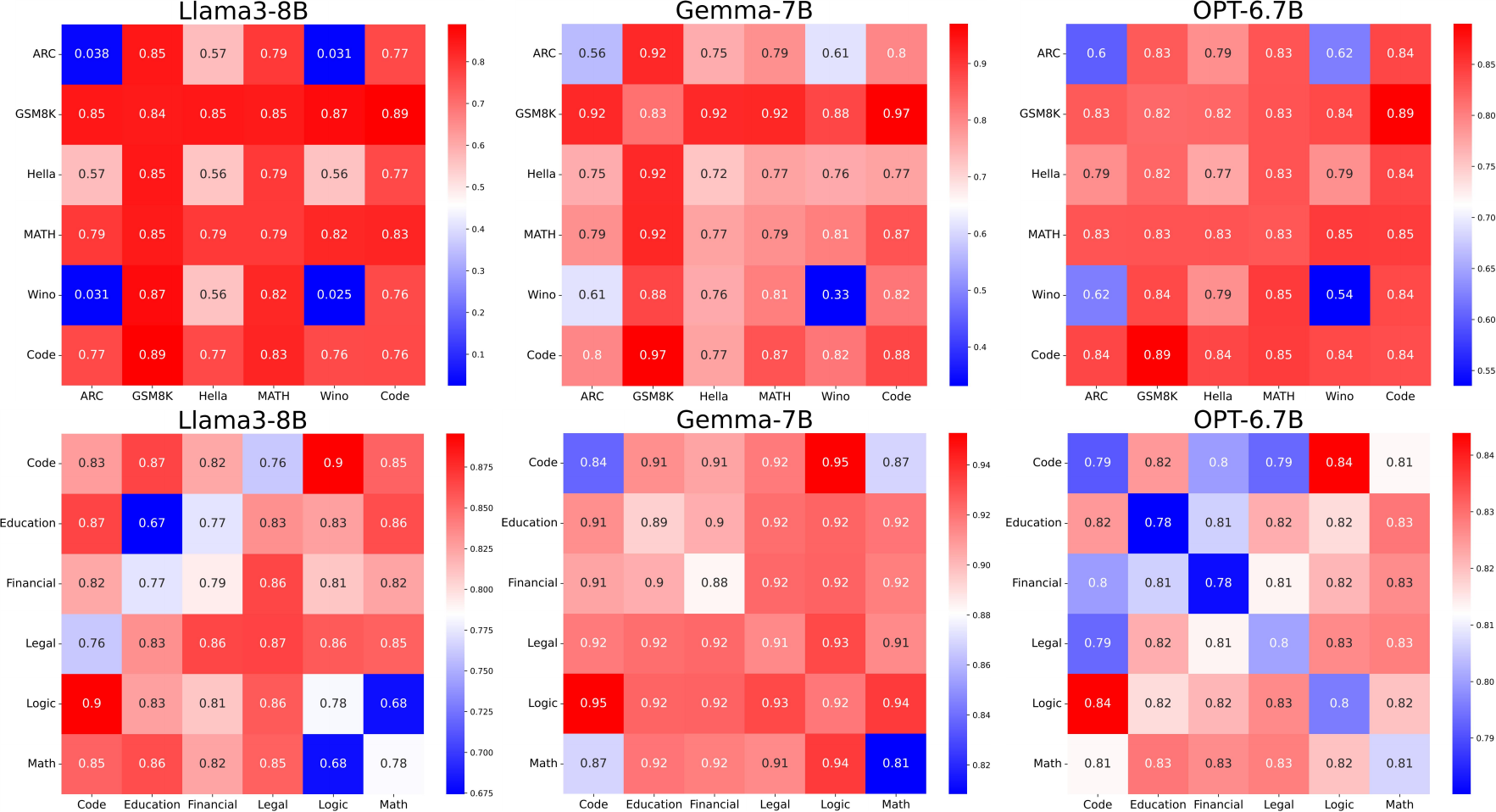}
\caption{\textbf{Top}: The least squares method was used to fit the activation pattern changes of traditional NLP tasks in SFT to the activation pattern changes in SGSM. A higher $R^2$ indicates a better fit, with Code Search Net and GSM8k showing the highest $R^2$ values. \textbf{Bottom}: The least squares method was used to fit the activation pattern changes of SFT instruction data to those of tasks requiring both “logical reasoning” skills and “programming and software development” instructions. The combination of “logical reasoning” skills and “programming” instructions achieved the highest $R^2$ value.}
\label{pic:r2}
\end{figure}
The upper part of Figure~\ref{pic:r2} shows the \(R^2\) values obtained by fitting the activation patterns of various traditional NLP tasks using the least squares method, with the activation pattern changes of the SGSM task as the dependent variable. Among these tasks, Code Search Net and GSM8K yield the highest \(R^2\) value of 0.97. This indicates that (1) the activation patterns of complex tasks can be well-fitted by the linear combination of activation patterns from simpler tasks; (2) SGSM is a task that involves solving elementary math problems using code, requiring a combination of elementary math and coding skills. In fact, the changes in SGSM’s activation patterns can be well-fitted by the activation pattern changes from the elementary math task GSM8K and the coding task dataset Code Search Net. 

A similar phenomenon can be observed in the results based on Infinity Instruct. The lower part of Figure~\ref{pic:r2} shows the \(R^2\) values obtained by fitting the activation patterns of various instruction-based NLP tasks using the least squares method, with the activation pattern changes of the instruction tasks that require both logical reasoning and programming/software development skills as the dependent variable. Among these tasks, the instructions that rely solely on logical reasoning and those that rely solely on programming/software development yield the highest \(R^2\) value of 0.95.

This shows that the activation patterns of complex tasks can be formed by combining the activation patterns of several simpler tasks, and this compositional relationship reflects the real-world connections between these tasks. Therefore, SFT adapts LLMs to complex tasks by integrating the attention heads corresponding to the simpler tasks.

Furthermore, the weigthed sum of activation patterns from models fine-tunned on simple tasks resembles that on complex tasks. It suggest that learned enough from basic tasks, its state will approximate that of a model directly fine-tuned with instructions related to complex tasks. This provides guidance on how to establish the necessary prerequisites for complex task performance.

\subsection{Small Data Can Change Activation Patterns}

In this section, we find that SFT can quickly adapt LLMs to certain basic tasks by adjusting the activation level of attention heads with rather limited instances. This suggests that with sufficient prior knowledge, the model can quickly adapt to a task even with a small amount of data.

\subsubsection{Analytical Method}

To analyze changes in instruction activation patterns during SFT, we saved checkpoints throughout the process for different tasks. We measured the similarity and distance between the activation patterns at different checkpoints using two metrics:  Correlation Coefficient and Mean Squared Error (MSE). Specifically, we calculated the Correlation and MSE between the activation patterns of the model at the \(i\)-th step of SFT on a given task and the activation patterns of the base model on the same task.
The Correlation reflects the consistency in activation pattern changes, while MSE captures the magnitude of these changes.

\subsubsection{Experiments and Results}

\begin{figure}[t]
\includegraphics[width=\linewidth]{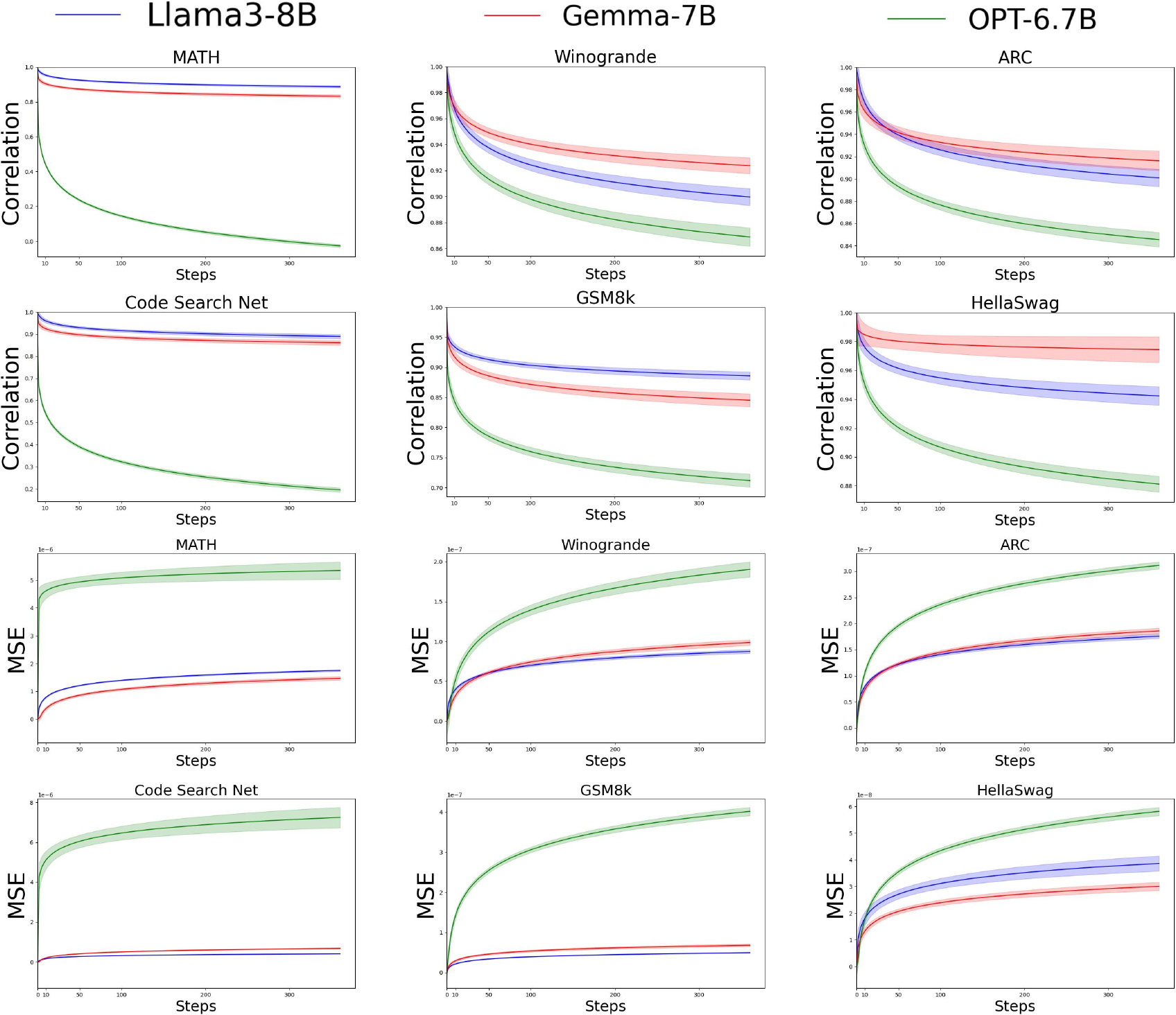}
\caption{Tracking changes in correlation coefficient
 and MSE activation patterns of the Llama3-8B, Gemma-7B, and OPT-6.7B models during fine-tuning on datasets including Code Search Net, GSM8k, MATH, SGSM, ARC, HellaSwag, and Winogrande.}
\label{fig:change_activation_patterns}
\end{figure}

As shown in Figure~\ref{fig:change_activation_patterns}, activation patterns change rapidly in the initial steps of SFT. This suggests that fine-tuning with small datasets can significantly reshape a model’s attention patterns and alter its performance. Previous research ~\cite{xia2024less} supports this, indicating that rapid learning during SFT is driven by swift changes in attention patterns.

Moreover, the extent of changes in activation patterns correlates with improvements in model performance. As shown in Figure~\ref{fig:change_activation_patterns}, the OPT-6.7B model shows significant changes in activation patterns across most tasks, particularly for complex tasks like MATH. The shifts in activation patterns post-fine-tuning are more pronounced compared to those in the Llama3-8B and Gemma-7B models. This suggests that the initial activation patterns of OPT-6.7B may not be well-suited for solving complex problems like MATH, and thus need more instances to approach the convergence. 

By analyzing models with varying capabilities, we find that for simpler tasks, fewer steps, samples, and less SFT data are needed for stronger LLMs to approach convergence and adapt to the task. In contrast, weaker models require more data due to having less prior knowledge from pre-training. This suggests that the foundation of rapid generalization is sufficient prior knowledge. Consequently, with enough prior knowledge, a small number of samples can enable rapid learning of complex tasks. Given that complex tasks can be decomposed into simpler ones, this raises the question: \textbf{Once a model has sufficient prior knowledge, can it be fine-tuned with just a small number of instructions to rapidly learn complex tasks?}

\section{Deduction and Applications}

We further validate our conclusions by testing their applicability in enhancing the effectiveness of SFT. Based on our observations of activation patterns, we focus on two scenarios: (1) Can we improve SFT on complex tasks by providing LLMs with prior knowledge of the basic tasks that make up the complex task? (2) Can we identify relevant data from a large candidate pool to approximate the effect of a target dataset, particularly when the target dataset is unavailable?

\subsection{Scenario 1: Combining Data of Basic Tasks to Promote the Learning of Complex Task}

Our analysis indicates that LLMs learn to complete complex tasks by leveraging a combination of their fundamental capabilities. A natural question arises: for a complex task, can we enhance the efficiency and effectiveness of the learning process by equipping LLMs with relevant basic capabilities? We conduct experiments on these settings: (1) Aiming at adapting LLMs to a complex task with a limited number of instructions given; (2) A large number of instructions about various basic tasks are available; (3) The relationship between the complex task and basic tasks are unknown.

Since the relationships between the complex task and basic tasks are unknown, we first estimate these relationships using the regression function described in Equation \ref{ercheng}. Intuitively, the regression coefficients \(\alpha_i\) describe the proportion that the complex task can be ``composed '' by a basic task $i$. Based on the regression coefficients, we create a preliminary instruction set by combining instructions from basic tasks.
\begin{equation}
\setlength{\abovedisplayskip}{-1pt}
\setlength{\belowdisplayskip}{-1pt}
    \text{Dataset}^{pre}=\{N \times \alpha_i / \sum_i \alpha_i \}_{i=0}^{|B|}
\end{equation}
where $|B|$ is the number of candidate basic tasks, $N$ is the predefined size of $\text{Dataset}^{pre}$. During experiment, LLM is first trained on $\text{Dataset}^{pre}$, then finetuned on the target task dataset $\text{Dataset}^{\text{complex}}$. 

\subsubsection{Experiment Settings}

To simulate scenarios where LLMs lack sufficient preliminary knowledge, we conducted experiments on the Llama series models, which have not been specifically optimized for Chinese corpora and exhibit relatively limited performance on Chinese-related tasks.
In this section, we set the target complex task as MathBench ~\cite{liu2024mathbench}, which requires the model to solve school mathematics questions using Chinese. The MathBench dataset 358 instances, and we use only 100 instances as $\text{Dataset}^{\text{complex}}$, the left part for testing. In addition to the basic tasks described in Section \ref{4.2}, we include RefGPT ~\cite{yang2023refgpt} (a dataset of Chinese factual knowledge) as a basic task dataset. After the regression process, we keep only basic datasets with the Top 2 largest regression coefficients, so as to filter out the noises. The remaining basic tasks are GSM8K (English elementary math) and RefGPT.Experiments are conducted on models of similar size but varying capabilities: Llama-7B ~\cite{touvron2023llama1}, Llama2-7B ~\cite{touvron2023llama2}, and Llama3-8B.

\begin{table}[t]
\centering
\small
\setlength{\belowcaptionskip}{-1pt}
\renewcommand{\arraystretch}{1.2}
\begin{tabular}{
    >{\centering\arraybackslash}m{1.6cm}
    >{\centering\arraybackslash}m{0.8cm}
    >{\centering\arraybackslash}m{0.8cm}
    >{\centering\arraybackslash}m{0.8cm}
    >{\centering\arraybackslash}m{0.8cm}
}
\toprule
\textbf{Model} & \textbf{Base}  & \textbf{SFT}  & \textbf{Random}  & \textbf{Ours}  \\ 
\midrule
\textbf{Llama-7B} & 28.68  & 31.78   & \underline{33.82}  & \textbf{36.82}  \\ 
\textbf{Llama2-7B} &29.07  & \underline{36.43}  & 34.51  & \textbf{40.70}  \\ 
\textbf{Llama3-8B} & 50.39   & \underline{52.33} & 51.16    & \textbf{55.03}  \\ 
\bottomrule
\end{tabular}
\caption{Performance on the Mathbench test set: "Base" refers to the base model results, ``SFT'' refers to results after fine-tuning with 100 Mathbench samples, and ``Random'' represents the average of five runs trained with a randomly mixed dataset of GSM8K and RefGPT data, followed by fine-tuning with 100 Mathbench samples. }
\label{tab:case1}
\end{table}

\subsubsection{Analysis}
As shown in Table \ref{tab:case1}, our approach consistently outperforms baseline models across all evaluations, including models fine-tuned solely on the Mathbench training set and those trained with a random mix of elementary mathematics and Chinese instruction data. Compared to models fine-tuned only on Mathbench, our method shows significant improvements in handling complex tasks through subtask training. Moreover, constructing datasets based on activation patterns improves the model's ability to leverage simple task data for complex tasks, surpassing models trained on randomly mixed subtasks.
For Llama-7B, learning from randomly proportioned basic task data can still improve complex task performance, but for more powerful models like Llama2-7B and Llama3-8B, this approach may hinder performance.
These findings suggest that guiding dataset construction using activation patterns from basic tasks significantly enhances the model's ability to learn and generalize to complex tasks, supporting our observation that the complex tasks could be the combination of basic complex tasks, and once equipping necessary preliminary knowledge the LLM could fast adapt to complex task.

\subsection{Scenario 2: Selecting Relevant Data from Candidates}

While various Domain-LLMs demonstrate strong capabilities, their instruction data is often unavailable. Considering the task-specificity of activation patterns, can we select relevant datasets from open instruction sets to approximate the performance of models trained on private domain datasets? Thus, we conduct experiments under the following settings: 
(1) only a small amount of developer-constructed pseudo-private data is available; (2) the complete dataset used to train the Domain-LLM is unknown;(3) the public data pool is large enough to contain data relevant to the target task.

\subsubsection{Experiment Settings}

\begin{table}[t]
\centering
\arraybackslash
\small
\setlength{\belowcaptionskip}{-1pt}
\renewcommand{\arraystretch}{1.2}
\begin{tabular}{
    >{\centering\arraybackslash}m{1.55cm}
    >{\centering\arraybackslash}m{0.35cm}
    >{\centering\arraybackslash}m{0.4cm}
    >{\centering\arraybackslash}m{0.4cm}
    >{\centering\arraybackslash}m{0.4cm}
    >{\centering\arraybackslash}m{0.4cm}
    >{\centering\arraybackslash}m{0.4cm}
    >{\centering\arraybackslash}m{0.4cm}
}
\toprule
\textbf{Model}& \textbf{Base}  & \textbf{T100}  & \textbf{R100}  & \textbf{T300} & \textbf{R300} &\textbf{T500}& \textbf{R500}\\ 
\midrule
\textbf{Llama-7B}& 30.00  & 31.66   & 30.33  & \underline{36.15 }&32.81 &\textbf{39.09 }&34.68  \\ 
\textbf{Llama2-7B}&35.08   & 39.05  & 37.06  & \underline{40.51 } &37.62 &\textbf{40.70}&37.95\\ 
\textbf{Llama3-8B}& 56.75   & 57.39&56.19&\textbf{57.95}&56.37&\underline{57.51}&56.19   \\ 
\bottomrule
\end{tabular}
\caption{Average accuracy of the model on four tasks: mathematics, physics, chemistry, and biology. "Base" represents the baseline model's average accuracy on these tasks. ``T100'' indicates the accuracy after fine-tuning with 100 most similar data points based on activation patterns. "R100" represents the accuracy after fine-tuning with 100 randomly selected unrelated MMLU data points, with similar meanings for ``T300'', ``R300'', ``T500'', and ``R500''.}
\label{tab:case2}
\end{table}
We simulated LLM performance without target data using the multi-domain MMLU dataset, designating test set data from specific domains as private (target data) and treating the remaining as non-target data. We used validation and dev set data from the target domains as pseudo-private data, simulating small, manually created datasets that developers might construct when original data is unavailable. 
As concluded in Section \ref{4.1}, fine-tuning on pseudo-private data improves the model’s task specificity in the target domain.

We used Equation \ref{case2} to identify the top $m$ non-target data points from MMLU most similar to the pseudo-private data. These data points were then used for further training. We compared the performance of this approach with the base model and models trained on an equal amount of randomly selected non-target data.

To ensure robustness, we averaged the results across multiple domains—mathematics, physics, chemistry, and biology—treating them as target data.
\begin{equation}
\setlength{\abovedisplayskip}{-1pt}
\setlength{\belowdisplayskip}{-1pt}
\label{case2}
\text{Dataset}^{app} = \operatorname{Top}_m  Corr(AP^{tar}, AP^{D_i}) ,
\end{equation}
where $\text{Dataset}^{app}$ represents the entire dataset, \(D_i\) is the \(i\)-th sample from \(D\), \(AP^{D_i}\) is the activation pattern for the \(i\)-th sample after SFT, and \(Corr(*,*)\) is the correlation coefficient.

\subsubsection{Analysis}

As shown in Table \ref{tab:case2}, models trained with data selected by activation pattern similarity consistently outperform those trained with randomly selected data. This suggests that using the similarity of activation patterns, we can effectively identify data that share similarities with the private data. These results support our hypothesis that attention heads carry task-specific information, enabling LLMs to adapt to various tasks. They also demonstrate the potential for identifying rare relevant data from large datasets to approximate private datasets. 


\section{Conclusion}
This study investigated the mechanisms behind the rapid learning and generalization observed during SFT. We found that attention heads are selectively activated in a task-specific manner during SFT, and that the activation patterns for complex tasks can be decomposed into combinations of basic task patterns. This offers a viable data substitution strategy when high-quality complex data is scarce.

Our experiments demonstrated that even small amounts of data can significantly alter activation patterns, validating the feasibility of (1) pre-training on simple tasks to equip large models with essential knowledge, and (2) fine-tuning with selected relevant data to enable rapid generalization. The improved effectiveness observed in these scenarios further supports the validity of our analytical approach and offers practical solutions to data challenges in complex and specialized tasks.

\section{Limitations}
While this study demonstrates the role of attention head activation patterns in rapid learning and generalization during the SFT process, we did not explore the specific impact of individual attention head activation levels on model performance in detail. Additionally, our method validation was primarily conducted on simpler tasks, and its applicability to more complex and real-world tasks remains to be fully evaluated. Future work will focus on finer-grained analysis of activation levels and further testing the effectiveness of our approach in more challenging and practical domains to enhance its performance in real-world applications.
\bibliography{custom}
\newpage
\appendix

\section{Dataset Details}
\label{Appendix I:Dataset Details}
In this paper, we utilize a wide range of datasets related to LLMs, many of which are commonly used in prominent large model evaluation frameworks \cite{open-llm-leaderboard-v2,2023opencompass}.

\begin{itemize}
\item MATH \cite{hendrycksmath2021}: An advanced mathematical reasoning dataset.
\item GSM8K \cite{cobbe2021gsm8k}: A dataset for elementary mathematical reasoning.
\item Code Search Net \cite{husain2019codesearchnet}: A large dataset of code sourced from GitHub; we specifically use the Python subset in this paper.
\item SGSM \cite{christ2024mathwell}: A dataset for solving math problems using code, requiring strong code generation and mathematical reasoning abilities.
\item HellaSwag \cite{zellers2019hellaswag}: A dataset for natural language processing and reasoning tasks.
\item Winogrande \cite{sakaguchi2021winogrande}: Another dataset for natural language processing and reasoning tasks.
\item ARC \cite{clark2018think}: A dataset for natural language processing and reasoning tasks.
\item Infinity Instruct dataset \cite{zhao2024iidoptimizinginstructionlearning}: A bilingual dialogue dataset; we use the English portion in this study. This dataset is annotated with the specific skills required to complete each dialogue, allowing us to determine which capabilities are needed for each task.
\item MathBench \cite{liu2024mathbench}: A dataset for evaluating mathematical performance in large models. We use the Chinese portions of the middle and primary subsets.
\item RefGPT \cite{yang2023refgpt}: A Chinese dataset focused on factual knowledge.
\item MMLU \cite{hendryckstest2021}: A dataset containing data from 57 domains. Following the approach of \citet{wang2024mmlupro}, we use "college mathematics," "abstract algebra," "elementary mathematics," "high school statistics," and "high school mathematics" to represent math-related data; "college physics," "high school physics," and "conceptual physics" for physics-related data; "college chemistry" and "high school chemistry" for chemistry-related data; and "college biology" and "high school biology" for biology-related data.
\end{itemize}
\section{Training Details}
\label{Appendix II:Training Details}
\begin{itemize}
    \item All experiments were conducted on an NVIDIA A100 GPU cluster.
    \item The training was performed using the DeepSpeed framework, following the hyperparameters outlined in \cite{yuan2023scaling}. We set ZeRO to 2, used a batch size of 4, a maximum token length of 1024, a learning rate of 1e-6, and gradient accumulation set to 4. Full fine-tuning was conducted with bf16 precision.
    \item Given the varying sizes of the datasets, to ensure consistency within each experiment, the number of training samples was set to match the smallest dataset in the group, ensuring that the results were not influenced by the amount of data.
    \item To ensure fair evaluation, all assessments were conducted using the OpenCompass framework \cite{2023opencompass}, employing greedy search to eliminate randomness in the results.
\end{itemize}

\section{Equation Details}

\label{Appendix III:Equation Details}

MSE represents the average distance between two matrices. The larger the MSE, the greater the difference between the two matrices.

\begin{equation}
\begin{split}
\text{MSE}(\mathbf{AP}^{\text{task1}},\mathbf{AP}^{\text{task2}}) 
&= \frac{1}{L \times H} \sum_{l=1}^{L} \sum_{h=1}^{H} \\
&\quad \left( \mathbf{AP}^{\text{task1}}_{l,h} - \mathbf{AP}^{\text{task2}}_{l,h} \right)^2,
\end{split}
\end{equation}

where \( \mathbf{AP}^{\text{task1}} \) and \( \mathbf{AP}^{\text{task2}} \) represent the activation pattern matrices for tasks 1 and 2, respectively, and \( L \) and \( H \) represent the number of layers and the number of heads per layer, respectively.

The Gini coefficient, which measures the inequality within the data of a matrix \( \mathbf{AP} \), can be expressed as:

\begin{equation}
G(\mathbf{AP}) = \frac{\sum_{i=1}^{L}\sum_{j=1}^{H}\sum_{k=1}^{L}\sum_{l=1}^{H} |\mathbf{AP}_{ij} - \mathbf{AP}_{kl}|}{2LH \sum_{i=1}^{L}\sum_{j=1}^{H} \mathbf{AP}_{ij}}
\end{equation}

Coefficient of Variation
The Coefficient of Variation (CV), which measures the relative variability of the data in matrix \( \mathbf{AP} \), is given by:

\begin{equation}
CV(\mathbf{AP}) = \frac{\sigma(\mathbf{AP})}{\mu(\mathbf{AP})},
\end{equation}

where \( \sigma(\mathbf{AP}) \) is the standard deviation of \( \mathbf{AP} \), and \( \mu(\mathbf{AP}) \) is the mean of \( \mathbf{AP} \).

The Kurtosis, which measures the peakedness of the data distribution in matrix \( \mathbf{AP} \), can be expressed as:

\begin{equation}
K(\mathbf{AP}) = \frac{LH \sum_{i=1}^{L}\sum_{j=1}^{H} (\mathbf{AP}_{ij} - \mu(\mathbf{AP}))^4}{\left( \sum_{i=1}^{L}\sum_{j=1}^{H} (\mathbf{AP}_{ij} - \mu(\mathbf{AP}))^2 \right)^2} ,
\end{equation}

where \( \mu(\mathbf{AP}) \) is the mean of the matrix \( \mathbf{AP} \). 

These expressions can be used in a LaTeX document to calculate the Gini coefficient, Coefficient of Variation, and Kurtosis for the matrix \( \mathbf{AP} \).

We use $R^2$ to analyze the task fitting performance:
\begin{equation}
\label{R2}
\small
R^2 = 1 - \frac{\sum_{l=1}^{L} \sum_{h=1}^{H} \left( \Delta AP^{\text{complex}}_{lh} - \Delta \hat{AP}^{\text{complex}}_{lh} \right)^2}{\sum_{l=1}^{L} \sum_{h=1}^{H} \left( \Delta AP^{\text{complex}}_{lh} - \overline{\Delta AP^{\text{complex}}} \right)^2}
\end{equation}

The coefficient of determination, denoted as \( R^2 \), is a statistical measure that indicates how well the data fits a regression model. In this context, \( \Delta AP^{\text{complex}}_{lh} \) represents the observed change in activation patterns at layer \( l \) and head \( h \) for a complex task, while \( \Delta \hat{AP}^{\text{complex}}_{lh} \) denotes the corresponding predicted change. The numerator captures the sum of squared errors between the observed and predicted values, whereas the denominator reflects the total variance in the observed data. \( \overline{\Delta AP^{\text{complex}}} \) represents the mean of the observed changes in activation patterns.

\end{document}